\documentclass{article} 
\usepackage{iclr2026_conference,times}
\iclrfinalcopy

\usepackage{amsmath,amsfonts,bm}

\def\eqref#1{equation~\ref{#1}}

\def\1{\bm{1}}

\DeclareMathAlphabet{\mathsfit}{\encodingdefault}{\sfdefault}{m}{sl}
\SetMathAlphabet{\mathsfit}{bold}{\encodingdefault}{\sfdefault}{bx}{n}

\usepackage{hyperref}
\usepackage{url}
\usepackage{tikz}
\usepackage{booktabs}
\usepackage{array}
\usepackage{multirow}
\usepackage{algorithm}
\usepackage{algpseudocode}
\usepackage{amsmath} 
\usepackage{microtype}
\usepackage{comment}
\usepackage{subcaption}
\usepackage{makecell}
\usepackage{setspace}

\usepackage{listings}
\usepackage{xcolor}

\usepackage{xcolor}
\usepackage{fancyvrb}
\usepackage{adjustbox}
\usetikzlibrary{positioning, calc}

\usepackage{algorithm}
\usepackage{algpseudocode}

\usepackage{algpseudocode} 

\algrenewcommand\algorithmicindent{0.8em}
\algrenewcommand\algorithmiccomment[1]{\hfill{\scriptsize$\triangleright$~#1}}

\definecolor{omittedcolor}{RGB}{180,30,30} 
\definecolor{bugcolor}{RGB}{30,90,180}     

\newcommand{\omitted}{\textcolor{omittedcolor}{// omitted}}

\newcommand{\bughl}[1]{\textcolor{bugcolor}{#1}}

\usetikzlibrary{arrows.meta, positioning, shapes.geometric, shapes.misc}

\title{Cross-Family Speculative Prefill: Training-Free Long-Context Compression with Small Draft Models}

\author{
Shubhangi Upasani\thanks{Corresponding author: \texttt{shubhangi.upasani@gmail.com}},
Ravi Shanker Raju,
Bo Li,
Mengmeng Ji,
John Long,
Chen Wu
\\
Urmish Thakker,
Guangtao Wang \\
SambaNova AI
}

\begin{document}

\maketitle

\begin{abstract}
Prompt length is a major bottleneck in agentic large language model (LLM) workloads, where repeated inference steps and multi-call loops incur substantial prefill cost. Recent work on speculative prefill demonstrates that attention based token importance estimation can enable training-free prompt compression, but this assumes the existence of a draft model that shares the same tokenizer as the target model. In practice however, agentic pipelines frequently employ models without any smaller in-family draft model. In this work, we study cross-family speculative prefill, where a lightweight draft model from one model family is used to perform prompt compression for a target model from a different family. Using the same speculative prefill mechanism as prior work, we evaluate a range of cross-family draft–target combinations, including Qwen, LLaMA, and DeepSeek models.

Across a broad diversity of tasks, we find that attention-based token importance estimation transfers reliably across different model families despite differences in model architectures and tokenizers between draft and target models. Cross-model prompt compression largely retains 90--100\% of full-prompt baseline performance and, in some cases, slightly improves accuracy due to denoising effects, while delivering substantial reductions in time-to-first-token (TTFT). These results suggest that speculative prefill depends mainly on task priors and semantic structure, thus serving as a generalizable prompt compression primitive. We discuss the implications of our findings for agentic systems, where repeated long-context inference and heterogeneous model stacks make cross-model prompt compression both necessary and practical.
\end{abstract}

\section{Introduction}

In agentic workloads, the performance of large language models (LLMs) is increasingly limited by the cost of prompt processing. Applications involving document-level parsing, tool use and code debugging have to make calls to a large language model i.e. the target model with long context that include reasoning traces, retrieved documents and execution logs. In these settings, the cost of prefill stage scales with prompt length and tends to become the primary bottleneck in terms of latency and throughput. 

Prompt compression therefore, is an attractive avenue to reduce LLM inference cost. Recent work on Speculative Prefill \citep{liu2025speculativeprefillturbochargingttft}  addresses this problem by estimating token-level importance scores using a lightweight draft model. In this setting, the target model is the large language model used to perform the final task inference on the compressed input prompt while the draft model is a smaller model used at inference time to estimate token importance to guide prompt compression. Both models belong to the same model family.  In practice however, some frontier models don't have smaller in-family draft models available (eg: DeepSeek-V3.1/R1, Kimi-K2 etc \citep{deepseek_v3.1_release_2025, Guo_2025, team2025kimi}). Moreover, agentic systems frequently combine heterogeneous models due to cost, availability, and deployment constraints. This raises a natural question: \emph{does attention-based token importance generalize across model families?} 

In this work, we show that attention-based speculative prefill generalizes across model families, and it can be used as a reliable prompt compression mechanism for target models which don't have a supporting in-family draft model. Importantly, we do not modify the original speculative prefill algorithm itself: we use the same attention-based token scoring and chunk selection as prior work \citep{liu2025speculativeprefillturbochargingttft}. The only difference is that the draft and target models are drawn from different model families, including combinations of Qwen, LLaMA, and DeepSeek models \citep{yang2025qwen3technicalreport, grattafiori2024llama3herdmodels,Guo_2025,liu2024deepseek}.

We summarize our key contributions below:
\begin{itemize}
    \item We present a simple, generalizable cross model approach to speculative prefill, showing that attention-driven token importance is largely transferable across model families.
    \item We evaluate our method on long context benchmarks like LongBench, RULER along with some code debugging tasks. Across a wide range of tasks, cross-model prompt compression largely retains 90--100\% of full-prompt baseline performance, despite substantial differences in model architecture and tokenizers between draft and target models\citep{bai2024longbenchbilingualmultitaskbenchmark,bai2025longbenchv2deeperunderstanding, hsieh2024rulerwhatsrealcontext, zhang2024infty}.
    \item We demonstrate up to a \(\sim\)18$\times$ reduction in time-to-first-token (TTFT), and show that cross-family prompt compression enables the practical use of long-context inputs under deployment constraints, by leveraging long-context draft models to compress inputs into a fixed deployable context budget for the target model.  
    \item Our findings have direct implications for agentic workloads, where prompt length can be long. Our method enables training-free, accuracy-preserving prompt compression even when the draft model is chosen purely for cost or availability reasons.

\end{itemize}


\section{Background}


Inference-time prompt compression techniques reduce the length of the input prompt at inference time, without modifying model parameters or requiring additional training. Existing approaches to inference-time prompt compression can be broadly categorized by how the prompt is reduced. Rewriting-based methods such as LLMLingua \citep{jiang2023llmlinguacompressingpromptsaccelerated} employ auxiliary models to remove less informative portions of the prompt, achieving substantial prompt compression. 
A second class of approaches performs token or chunk-level selection, directly retaining a subset of tokens from the original prompt while preserving their original ordering. Recent work on attention-based token selection, most notably speculative prefill \citep{liu2025speculativeprefillturbochargingttft}, instantiates this paradigm by using internal model signals to identify salient tokens at inference time. 

Alongside prompt compression, a large body of work focuses on orthogonal inference optimizations that do not modify the input prompt itself. Attention acceleration techniques such as FlashAttention, sparse  attention mechanisms (e.g., MInference, DuoAttention, LongFormer), and KV cache management strategies (e.g., H2O, SnapKV, SageKV, StreamingLLM) improve inference efficiency by reducing computation or memory usage during prefill and decoding respectively  \citep{xiao2024duoattentionefficientlongcontextllm,jiang2024minference10acceleratingprefilling,dao2022flashattentionfastmemoryefficientexact,beltagy2020longformerlongdocumenttransformer,zhang2023h2oheavyhitteroracleefficient,li2024snapkvllmknowslooking,wang2025llmsknowdropselfattention,xiao2024efficientstreaminglanguagemodels}. While complementary to prompt compression, these methods operate after the prompt has already been ingested by the model.

\begin{figure}[t]
\centering
\scriptsize

\begin{tikzpicture}[
    node distance=8mm and 8mm,
    box/.style={draw, rounded corners=2pt, align=center,
                minimum width=26mm, minimum height=6mm},
    highlight/.style={draw=red!70!black, rounded corners=3pt,
                      align=center, fill=yellow!10,
                      minimum width=30mm, minimum height=7mm},
    arrow/.style={->, thick}
]

\node[box, fill=orange!20] (input)
{Long Input Prompt};

\node[box, fill=violet!18, right=of input] (draft)
{Small Draft LLM \\ (cross-family)};

\node[box, draw=red!70!black, fill=yellow!15, right=of draft] (attn)
{Attention Importance Estimator \\ token/chunk saliency from \\ draft attention};

\node[highlight, right=of attn] (select)
{Select Top-$K$ blocks\\Concat selected text blocks};

\node[box, fill=orange!20, below=of select] (reduced)
{Reduced (text) input};

\node[box, fill=orange!20, left=of reduced] (tokenizer)
{Target LLM Tokenizer};

\node[box, fill=green!30, left=of tokenizer] (target)
{Target LLM Inference};

\node[box, fill=green!18, left=of target] (resp)
{Response};

\draw[arrow] (input) -- (draft);
\draw[arrow] (draft) -- (attn);
\draw[arrow] (attn) -- (select);

\draw[arrow] (select) -- (reduced);

\draw[arrow] (reduced) -- (tokenizer);
\draw[arrow] (tokenizer) -- (target);
\draw[arrow] (target) -- (resp);

\end{tikzpicture}

\caption{Cross-model speculative prefill prompt compression pipeline}
\label{fig:specprefill_wrapped}
\end{figure}
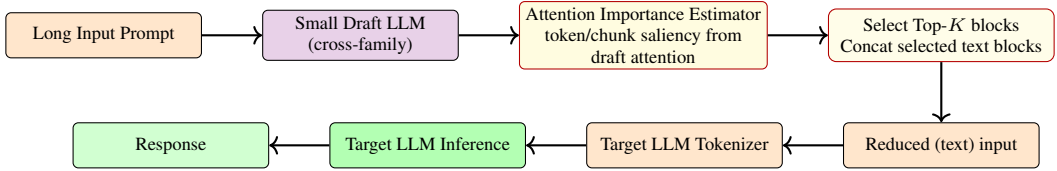











\section{Method}

We adopt the speculative prefill architecture and token selection mechanism introduced in prior work, using a lightweight draft model from a different model family to identify salient tokens in the input prompt as shown in Figure \ref{fig:specprefill_wrapped}. Our method follows speculative prefill’s core algorithmic steps but with a small but important modification to support heterogeneous draft–target model pairs. We use the same attention-based token importance estimation and chunk-based selection procedure as prior work, and defer details to Appendix~\ref{app:additional_method_details}. Below, we describe the modifications required to enable cross-family compatibility.

\subsection{Keep Rate Selection}

We define the keep rate $\rho$ with respect to the target model’s tokenization. Given an original target-side prompt length $L_{\text{target}}$ and a desired compressed length $\hat{L}_{\text{target}}$, we set $\rho = \hat{L}_{\text{target}} / L_{\text{target}}$. This rate is applied in the draft domain by selecting the top-$K$ chunks corresponding to a fraction $\rho$ of the draft sequence based on attention-derived importance scores. The selected spans are converted back to text and re-tokenized using the target tokenizer to form the compressed prompt. Due to cross-family tokenization differences, the resulting target-side length may deviate slightly from $\hat{L}_{\text{target}}$, but is typically close in practice. Since $L_{\text{target}}$ varies across inputs, $\rho$ is computed per sample, and we report its average value for each benchmark.

For most of our results reported in Section 5, we select keep rates that correspond to compressed prompt lengths of 8k, 16k, 24k, and 32k tokens, which are compatible with the paged-attention execution scheme on our custom hardware Reconfigurable Dataflow Unit (RDU). In particular, KV caches are stored in 4k-token blocks, and attention computation operates over these blocks during inference. Choosing compressed prompt lengths that are multiples of this block size ensures efficient KV cache paging and streaming during attention computation.

\subsection{Cross-Family Draft--Target Compatibility}

After chunk selection in the draft domain, adjacent chunks are merged into contiguous spans and concatenated at the text level to construct the compressed prompt. Task-specific delimiter tokens are inserted between non-contiguous spans to explicitly mark discontinuities. Unlike the original speculative prefill formulation, which restores original positional indices, we assign new, contiguous position IDs before passing the compressed prompt to the target model (refer Algorithm \ref{alg:cross_family_specprefill}). This avoids the need for position alignment across heterogeneous tokenization schemes, and we find that it has negligible impact on task accuracy across all evaluated benchmarks.

\begin{algorithm}[t]
\caption{Cross-Family Speculative Prefill (Prompt Compression)}
\label{alg:cross_family_specprefill}
{\fontsize{8.3pt}{8pt}\selectfont
\begin{algorithmic}[1]
\Require Target model $T$, draft model $D$, prompt text $x$, lookahead steps $N$, keep rate $\rho$, chunk size $c$, delimiter token $\langle\mathrm{DELIM}\rangle$
\Ensure Target output $y$

\State $\mathcal{A} \gets \emptyset$
\State $t \gets \Call{Tokenizer}{D, x}$ \Comment{Draft-tokenize; length $S$}
\For{$i = 1$ to $N$}
    \State $(\mathrm{attn}_i, \_) \gets \Call{ForwardWithAttn}{D, t, \mathrm{lookahead}=i}$
    \State $\mathcal{A} \gets \mathcal{A} \cup \{\mathrm{attn}_i\}$
\EndFor
\Comment{$\mathrm{attn}_i \in \mathbb{R}^{L \times H \times S}$}

\State $I \gets \Call{AggregateMaxMean}{\mathcal{A}}$ \Comment{$I \in \mathbb{R}^{S}$}
\State $\tilde{I} \gets \Call{AvgPool1D}{I}$
\State $\mathcal{C} \gets \Call{PartitionIntoChunks}{\tilde{I}, c}$
\State $\mathcal{J} \gets \Call{TopKChunks}{\mathcal{C}, \rho}$
\State $\mathcal{S} \gets \Call{MergeAdjacentChunks}{\{\mathcal{C}_j : j \in \mathcal{J}\}}$

\State $\mathcal{M} \gets \Call{MapDraftChunksToTextSpans}{t, \mathcal{S}}$
\State $x_{\mathrm{cmp}} \gets \Call{ConcatWithDelim}{\mathcal{M}, \langle\mathrm{DELIM}\rangle}$

\State $u \gets \Call{Tokenizer}{T, x_{\mathrm{cmp}}}$ \Comment{Target-tokenize}
\State $p \gets (0,1,\ldots,|u|-1)$ \Comment{New contiguous pos-ids}
\State $y \gets \Call{Forward}{T, u, p}$
\State \Return $y$
\end{algorithmic}
}
\end{algorithm}

\section{Experimental Setup}

\subsection{Benchmarks}

We evaluate cross-family speculative prefill on a suite of long-context benchmarks covering diverse input structures and reasoning requirements, including LongBench v1 and v2, RULER, and Code Debug task from InfiniteBench~\citep{bai2024longbenchbilingualmultitaskbenchmark,bai2025longbenchv2deeperunderstanding,hsieh2024rulerwhatsrealcontext,zhang-etal-2024-bench}. LongBench v1 and v2 assess multi-document reasoning, retrieval, summarization, and code understanding across varying context lengths, while RULER and InfiniteBench stress extreme long-context settings such as needle-in-a-haystack retrieval and repository-scale code debugging. We present additional details in Appendix \ref{app:benchmark_details}. 

\subsection{Models and Draft--Target Pairings}

We evaluate cross-family speculative prefill using a diverse set of heterogeneous target--draft model pairings spanning the Qwen, LLaMA, and DeepSeek model families. On LongBench v1 and v2, we use mid-scale target models (e.g., Qwen-8B and LLaMA-8B) paired with smaller draft models, and additionally evaluate DeepSeek-R1 as a target on LongBench v2. To study extreme long-context settings, we use RULER with DeepSeek-V3 as the target model. For code debugging tasks, we evaluate both DeepSeek-R1 and DeepSeek-V3.1 as target models.  For each pairing, we explore a range of keep rates and report accuracy relative to the corresponding full-prompt baseline. Keep rates reported in the results denote average values across each benchmark, while the target-side compressed context length is held fixed for all samples within a given target--draft pairing. Additional implementation details are provided in Appendix~\ref{app:impl_details}.

\section{Results}

\subsection{Cross-Family Speculative Prefill Preserves Task Performance}


Results for LongBench v2 and Code Debug are presented in Tables~\ref{tab:long_bench_v2} and~\ref{tab:code_debug}, with additional results on LongBench v1 and RULER reported in Tables~\ref{tab:long_bench_v1} and \ref{tab:ruler} respectively. We use greedy decoding for all reported results. For Code Debug, the average input length for DeepSeek-R1 is \(\sim\)110k tokens, and keep rates of 30\%, 20\%, and 15\% reduce the effective context to roughly 33k, 22k, and 16k tokens respectively.  For LongBench v2, the average input length is 242k, 260k and 248k tokens under LLaMA-8B, Qwen3-8B and DeepSeek-R1 as target models, and the compressed lengths follow the keep rates specified in Table~\ref{tab:long_bench_v2}.

Across diverse benchmarks, cross-model speculative prefill maintains performance close to the full-prompt baseline (typically within 90–100\%), even at low keep rates of 6\% on selected Longbench V2 tasks. This robustness holds despite differences in architecture and tokenization between the draft and target model families, suggesting that token importance estimates are not strongly tied to a specific model family. 
For LLaMA-8B, we evaluate both an in-family draft (LLaMA-1B) and a cross-family draft (Qwen3-1.7B). The results show comparable performance across keep rates (e.g., 29.4\% vs 29.6\% at 25\% keep rate, and 30.6\% vs 31.2\% at 50\%). Similar trends are observed for Qwen3-8B when paired with LLaMA-1B and Qwen3-1.7B drafts. These results indicate that speculative prefill is not strongly dependent on cross-family pairing, and that small draft models of comparable size provide similar token saliency signals for prompt compression. 

For DeepSeek-R1, we additionally explore keep rates of 3\%, 6\%, and 10\% on LongBench-v2. These correspond to compressed prompt lengths of approximately 8k, 16k, and 24k tokens, respectively. We observe that performance improves as the keep rate increases from 3\% to 6\%, indicating the expected trade-off between compression and task performance. The accuracy decreases slightly at 10\%, suggesting a diminishing return from retaining additional context. Finally, we note that the larger draft model (Qwen3-4B-Instruct-2507) achieves higher performance on DeepSeek-R1 because it supports a 262k native context length, compared to 32k for Qwen3-1.7B. The longer context allows the draft model to estimate attention over long inputs more reliably, which improves token selection during compression.

We see some degradation on Code Debug under aggressive compression (e.g., 15\% keep rate), which is discussed in Appendix~\ref{app:limitations}. In some cases, most notably on LongBench v2, compressed prompts achieve performance slightly exceeding the full-prompt baseline (Table ~\ref{tab:long_bench_v2}). We attribute this to the denoising effect of context compression: by removing large amounts of irrelevant context, speculative prefill simplifies the problem.

For RULER with input context lengths of 128k (Table ~\ref{tab:ruler}), we note that the full-prompt baseline is evaluated using SnapStream KV cache compression \citep{li2025snapstream} on our custom RDU hardware. This is needed to make 128k-token inference feasible within on-chip memory constraints for DeepSeek-V3. This cache compression contributes to loss of long range information needed for retrieval-heavy tasks in RULER, thus observing the lower baseline performance.  In contrast, speculative prefill reduces the effective context length prior to inference, thus proactively removing noisy context and mitigating this degradation. Together, these results highlight that cross model speculative prefill can preserve accuracy and in some cases, improve task performance under practical deployment constraints. Overall, cross-model speculative prefill enables substantial prompt compression while largely preserving downstream task performance. Additional details are given in \ref{app:more_results}.

\subsection{Cross model prompt compression reduces long context cost}

On RULER (Table~\ref{tab:ruler}), cross-model speculative prefill yields substantial latency improvements by aggressively compressing long inputs. Compressing 128k token prompts to 16k using a lightweight draft model reduces TTFT from 46 seconds to approximately 2.5 seconds, corresponding to a \(\sim\)18$\times$ reduction. Even at a more conservative compression target of 32k tokens, TTFT is reduced to 4.3 seconds. We measure these latencies on our custom RDU hardware using the same TTFT measurement methodology commonly used for GPU-based systems.

\begin{table*}[t]
\centering
\begin{minipage}[t]{0.52\textwidth}
\centering

\caption{Cross-model speculative prefill on LongBench V2 (\textbf{SP} = speculative prefill)}

\label{tab:long_bench_v2}
{\scriptsize
\setlength{\tabcolsep}{2.5pt}
\renewcommand{\arraystretch}{0.95}
\renewcommand\cellgape{\Gape[1pt]}

\begin{tabular}{@{}>{\raggedright\arraybackslash}p{2.6cm}ccccccc@{}}
\toprule
Method & Keep rate & easy & hard & short & medium & long & \makecell[c]{Accuracy \\ (\%)} \\ 
\midrule
\addlinespace[1pt]
\multicolumn{8}{c}{\bfseries Target -- Llama-3.1-8B-Instruct} \\
\addlinespace[1pt]
Full Prompt  & -- & 31.2 & 31.2 & 36.1 & 29.3 & 26.9 & 31.2 \\ 
\makecell[l]{SP w/ Llama-3.2-1B-Ins} & 25\% & 30.2 & 28.9 & 39.4 & 23.7 & 24.1 & 29.4 \\
\makecell[l]{SP w/ Qwen3-1.7B} &  25\% & 32.8 & 27.7 & 38.3 & 26.5 & 21.3 & 29.6\\
\makecell[l]{SP w/ Llama-3.2-1B-Ins} & 50\% & 32.8 & 29.3 & 40 & 26.5 & 23.1 & 30.6 \\
\makecell[l]{SP w/ Qwen3-1.7B} & 50\% & 31.8 & 30.9 & 37.8 & 28.4 & 25.9 & 31.2\\
\makecell[l]{SP w/ Qwen3-0.6B} & 10\% & 33.3 & 28.0 & 35.6 & 28.8 & 23.1 & 30.0\\

\midrule
\addlinespace[1pt]
\multicolumn{8}{c}{\bfseries Target -- Qwen3-8B}\\
\addlinespace[1pt]
Full Prompt  & -- & 30.7 & 28.3 & 39.4 & 23.7 & 23.1 & 29.2 \\
\makecell[l]{SP w/ Qwen3-1.7B} & 25\% & 35.4 & 29.6 & 37.8 & 27.9 & 29.6 & 31.8 \\
\makecell[l]{SP w/ Llama-3.2-1B-Ins}  & 25\% & 32.3 & 26.0 & 34.4 & 24.2 & 26.9 & 28.4 \\ 
\makecell[l]{SP w/ Qwen3-1.7B} & 50\% & 33.9 & 26.7 & 38.9 & 22.3 & 27.8 & 29.4 \\
\makecell[l]{SP w/ Llama-3.2-1B-Ins} & 50\% & 35.9 & 26.4 & 39.4 & 25.1 & 24.1 & 30.0 \\ 

\midrule
\addlinespace[1pt]
\multicolumn{8}{c}{\bfseries Target -- DeepSeek-R1}\\
\addlinespace[1pt]
Full Prompt  & -- & 66.1 & 53.4 & 62.2 & 54.4 & 59.3 & 58.3 \\
\makecell[l]{SP w/ Qwen3-4B-Ins-2507}  & 6\% & 57.8 & 50.5 & 64.4 & 45.1 & 50.9 & 53.3 \\ 
\makecell[l]{SP w/ Llama-3.1-8B-Ins}  & 6\% & 55.2 & 53.4 & 60.6 & 45.6 & 60.2 & 54.1 \\ 
\makecell[l]{SP w/ Qwen3-1.7B} & 3\% & 50 & 43.4 & 50.6 & 38.1 & 53.7 & 45.9 \\
\makecell[l]{SP w/ Qwen3-1.7B} & 6\% & 54.2 & 44.1 & 56.1 & 39.5 & 50.9 & 47.9 \\
\makecell[l]{SP w/ Qwen3-1.7B} & 10\% & 53.6 & 42.8 & 58.3 & 37.7 & 46.3 & 46.9 \\
\bottomrule
\end{tabular}
}
\end{minipage}
\hfill
\begin{minipage}[t]{0.4\textwidth}
\centering
\caption{Cross-model speculative prefill on Code Debug}
\label{tab:code_debug}
{\scriptsize
\setlength{\tabcolsep}{2.5pt}
\renewcommand{\arraystretch}{0.95}
\renewcommand\cellgape{\Gape[1pt]}

\begin{tabular}{@{}>{\raggedright\arraybackslash}p{2.3cm}cc@{}}
\toprule
Method & Keep rate & \makecell[c]{Accuracy \\ (\%)} \\ 
\midrule
\addlinespace[1pt]
\multicolumn{3}{c}{\bfseries Target -- DeepSeek-V3.1} \\
\addlinespace[1pt]
Full Prompt  & -- & 67.51 \\ 
\makecell[l]{SP w/\\Llama-3.1-8B-Ins}  & 20\% & 64.72 \\
\makecell[l]{SP w/\\Llama-3.1-8B-Ins}  & 15\% & 59.13 \\

\midrule
\addlinespace[1pt]
\multicolumn{3}{c}{\bfseries Target -- DeepSeek-R1-0528}\\
\addlinespace[1pt]
Full Prompt  & -- & 74.37 \\
\makecell[l]{SP w/\\Llama-3.1-8B-Ins}  & 30\% & 70.30 \\
\makecell[l]{SP w/\\Llama-3.1-8B-Ins}  & 25\% & 68.02 \\ 
\makecell[l]{SP w/\\Llama-3.1-8B-Ins}  & 15\% & 62.44 \\ 
\bottomrule
\end{tabular}
}
\end{minipage}
\end{table*}

\begin{table}[h!]
\centering
\caption{Cross-model speculative prefill results on LongBench V1}
\label{tab:long_bench_v1}
\resizebox{\textwidth}{!}{%
\begin{tabular}{@{}lcccccccccc@{}}
\toprule
Method & Keep rate & hotpotQA & pass.ret.en & qmsum & mulfqa.en & lcc  & rep-p & mulfqa.zh & pass.ret.zh & \makecell[c]{Accuracy \\ (\%)} \\ \midrule
\multicolumn{11}{c}{\textbf{Target - Llama3.1-8B-Instruct}} \\
\midrule
Full Prompt  & - & 55.97 & 99.5 & 25.38 & 55.8 & 44.56 & 43.5 & 62.45 & 97.38 & 55.31 \\ 

Spec prefill with Qwen3-1.7B & 45\% & 55.99 & 100 & 24.9 & 54.32 & 41.09 & 41.33 & 62.4 & 95.25 & 54.29  \\

\midrule
\multicolumn{11}{c}{\textbf{Target-Qwen3-8B}}\\
\midrule
Full Prompt  & - & 59.47 & 100 & 24.03 & 53.84 & 50.49 & 51.13 & 62.45 & 98.5 & 57.34\\

Spec prefill with 
Llama-3.2-1B-Instruct  & 45\% & 59.63 & 100 & 24.05 & 52.56 & 46.21 & 49.74 & 59.69 & 99 & 55.98 \\

 \bottomrule
\end{tabular}%
}
\end{table}

\begin{table}[h!]
\centering
\caption{Cross-model speculative prefill results on RULER}
\label{tab:ruler}
\resizebox{\textwidth}{!}{%
\begin{tabular}{@{}lcccccccccccc@{}}
\toprule
Method & Keep rate & niah\_single\_1 & niah\_single\_2 & niah\_single\_3 & niah\_multikey\_1 & niah\_multikey\_2  & niah\_multikey\_3 & niah\_mutlivalue & niah\_multiquery & qa\_1 & qa\_2 & \makecell[c]{Accuracy \\ (\%)}\\ \midrule
\multicolumn{13}{c}{\textbf{Target - DeepSeek V3}} \\
\midrule
Full Prompt  & - &  100 & 100 & 62.8 & 89.2 & 99.6 & 32.2 & 98.75 & 99.25 & 61.2 & 55 & 80\\ 

Spec prefill with  \\ Qwen3-4B-Ins-2507 & 12.5\% &  100 & 100 & 100 & 100 & 83.4 & 76.4 & 99.6 & 100 & 70.46 & 66.84 & 89.67\\

Spec prefill with \\ Llama-3.1-8B-Instruct & 12.5\% &   100 & 100 & 100 & 99.8 & 99.6 & 96.8 & 99.2 & 100 & 70.74 & 67.43 & 93.36\\

 \bottomrule
\end{tabular}%
}
\end{table}


\subsection{Robustness to Draft Model Selection}

We next evaluate the robustness of prompt compression to the choice of draft model while holding the target model fixed. Results on LongBench v2 (Table~\ref{tab:long_bench_v2}) indicate that effective compression can be achieved using draft models with substantially different parameter scales. For instance, with LLaMA-8B as the target model, both Qwen3-1.7B and Qwen3-0.6B produce compressed prompts that preserve performance close to the full-prompt baseline. Likewise for DeepSeek-R1, competitive results are obtained using either Qwen3-4B or LLaMA-8B as draft models. Our experiments show that small draft models (1B–1.7B range) already provide sufficiently reliable saliency estimates for 8B target models. Larger draft models (e.g., Qwen3-4B) improve performance primarily when processing very long inputs, where stronger long-context modeling leads to more accurate attention estimation. Overall, performance varies only modestly across draft model choices, suggesting that the compression quality is not strongly dependent on a specific architecture or model scale beyond a minimal effectiveness threshold. Tokens that are important for reasoning (numbers, entities, key instructions) tend to attract attention consistently across different LLM architectures trained on similar data. This robustness enables flexible draft model selection based on computational cost or resource constraints without sacrificing downstream accuracy.


\subsection{Decoupling Target Context Length from Deployment Limits}

Although recent open-source models such as DeepSeek-V3.1/R1 support contexts up to 128k tokens, deploying them at full length is often infeasible due to memory and hardware constraints, resulting in much smaller usable context windows in practice.  For our RDU, we were limited to a maximum target side context length of 32k tokens for these models. Cross-family speculative prefill enables long-context capability by using small draft models with native long-context support (e.g., LLaMA-8B and Qwen3-4B) to process long prompts and compress them via attention-based token importance before forwarding them to these target models within a 32k-token deployment budget.  This allows target models to effectively operate on information from 128k-token inputs while maintaining performance close to the full-prompt baseline on Code Debug (Tables~\ref{tab:code_debug}).  More generally, the same mechanism enables scalable context extension: a target model with limited context capacity (e.g., 128k) can leverage a longer-context draft model (e.g., 256k) to compress longer inputs and simulate full-context inference beyond its native deployment limit.  


\section{Conclusion}

Prior speculative prefill assumes in-family drafts; we show this assumption is unnecessary. We studied cross-family speculative prefill, where a lightweight draft model from one model family estimates attention-based token importance to compress prompts for a target model from a different family. Without modifying speculative prefill’s core algorithm, we support heterogeneous draft--target pairs. Across LongBench, RULER, and code debug, our results show that attention-derived importance signals are largely transferable across model families with some task-specific constraints (Appendix \ref{app:limitations}), enabling substantial prompt compression while preserving most of the full prompt accuracy and lending significant latency benefits through TTFT reduction. Crucially, this flexibility enables practical long-context deployment under real-world hardware constraints, by allowing any available long-context draft model to compress inputs for a constrained target model. Our results have direct implications for agentic workloads where prompt processing is a major bottleneck.


\newpage

\bibliography{iclr2026_conference}
\bibliographystyle{iclr2026_conference}

\appendix
\section{Appendix}

\subsection{Token Importance Estimation and Chunk Selection}
\label{app:additional_method_details}
\subsubsection{Attention-Based Token Importance Estimation}

Given an input prompt of length S, the draft model processes the full context and computes attention scores using a fixed number of lookahead tokens. As in prior work \citep{liu2025speculativeprefillturbochargingttft}, lookahead is used to mitigate positional biases in attention distributions by aggregating signals from multiple decoding steps. Let the attention tensor be of shape \textit{[N, L, S, H]}, where \textit{N} is the number of lookahead tokens, \textit{L} the number of transformer layers, \textit{S} the prompt length, and \textit{H} the number of attention heads. We compute token importance scores using the same max–mean aggregation strategy as speculative prefill: attention scores are first maximized over the layer and head dimensions, and then averaged across the lookahead dimension. This produces a single scalar importance score for each input token.

\subsubsection{Chunk-Based Prompt Compression}

To reduce variance in token-level importance estimates, we perform prompt compression using a chunk-based selection strategy identical to speculative prefill. We first apply one-dimensional average pooling over the token importance scores. The input prompt is then partitioned into contiguous chunks, and the average importance score within each chunk is computed. Finally, the Top-K chunks are selected based on these average scores, yielding a compressed prompt.

\subsection{Benchmark details}
\label{app:benchmark_details}

\paragraph{LongBench v1 and v2:}
We evaluate eight representative tasks spanning multi-document question answering (HotpotQA), query-based summarization (QMSum), passage retrieval (PassageRetrieval), long-form code understanding (LCC), and repository-level code reasoning (RepoBench)~\citep{yang2018hotpotqadatasetdiverseexplainable,zhong2021qmsumnewbenchmarkquerybased,guo2023longcoderlongrangepretrainedlanguage,liu2023repobenchbenchmarkingrepositorylevelcode}. We use the full evaluation set for LongBench v2, which includes substantially longer and more diverse inputs than LongBench v1.


\paragraph{RULER and InfiniteBench:}
RULER focuses on extreme long-context reasoning tasks such as Needle-in-a-Haystack and long-context QA, while Code.Debug task from InfiniteBench requires identifying injected errors within large code repositories.

\subsection{Implementation Details}
\label{app:impl_details}

For all experiments, we use greedy decoding for answer generation. Discontinuous text spans produced by draft-model compression are separated using task-specific delimiter tokens. For LongBench and RULER results, we used \texttt{[...]} as the delimeter token. We vary the prompt keep rate across benchmarks while fixing the number of lookahead tokens to \(N = 8\). For LongBench v1/v2 and RULER, we use a chunk size of 32 for top-\(K\) selection and apply one-dimensional average pooling with a kernel size of 13 to smooth attention scores prior to chunk selection.

For the code debugging task from InfiniteBench, we retain the above settings with two modifications. First, we use the delimiter token ``// omitted'' to explicitly mark non-contiguous code regions. Second, we increase the chunk size to 128 to better align with the structure of large code repositories. In addition, we make two evaluation-related adjustments to the InfiniteBench codebase: (i) we extend the answer extraction logic to include additional matching patterns to avoid false negatives, and (ii) we remove explicit \texttt{<think>} tags from DeepSeek-R1 responses prior to evaluation, which we find improves accuracy by preventing spurious mismatches during answer parsing. These adjustments apply equally to full prompt results also to make them directly comparable. 


\subsection{Additional result Analysis}
\label{app:more_results}

We report results on LongBench v1 and RULER in Tables~\ref{tab:long_bench_v1} and~\ref{tab:ruler}. On LongBench v1, we apply cross-model speculative prefill to compress inputs to a fixed target length of 4k tokens for each target model, corresponding to an average keep rate of 45\%. Across the selected LongBench v1 datasets, the average input length is 8805 tokens for LLaMA-8B and 8867 tokens for Qwen3-8B as target models. Draft models from both the Qwen and LLaMA families act as effective compressors, with cross-family prompt compression achieving accuracies close to the full-prompt baseline.

For RULER, we compress 128k-token inputs down to 16k tokens (12.5\% keep rate). Notably, the full-prompt baseline underperforms speculative prefill in this setting. We attribute this to two main factors. First, speculative prefill performs implicit context denoising by retaining only salient regions of the input, which improves accuracy on retrieval-heavy, needle-in-a-haystack tasks. This denoising is missing for the reported baseline results. Second, making 128k full-prompt inference feasible with DeepSeek V3 requires combining it with SnapStream KV cache compression \citep{li2025snapstream} in our deployment setting. While necessary for memory reasons, KV cache compression can discard long-range information, leading to degradation in the full-prompt baseline, particularly for RULER which has retrieval heavy tasks.

In contrast, speculative prefill enables effective 128k input processing for DeepSeek V3, which is otherwise limited to a 32k context window on our hardware. By leveraging draft models (eg: LLaMA-8B) with native 128k context support, cross-family speculative prefill extends the usable context length of the target model from 32k to 128k, enabling long-context inference that would otherwise be infeasible under deployment constraints. The performance with speculative prefill is also much better than the baseline results we get with SnapStream KV cache compression. These results highlight the practical benefit of speculative prefill in enabling long-context inference for high-capacity target models, while simultaneously improving performance through token-aware prompt compression.

\subsection{Future research directions}
\label{app:limitations}
While cross-family speculative prefill preserves near-baseline performance on most benchmarks, we observe a clear limitation on long-context code debugging. In Table \ref{tab:code_debug}, aggressive prompt compression (e.g., a 15\% keep rate) leads to a noticeable accuracy drop for both DeepSeek-V3.1 and DeepSeek-R1. Specifically, DeepSeek-V3.1 decreases from 67.51\% (full prompt) to 59.13 \% at 15\% keep rate, retaining $\sim$87.6\% of baseline performance, and DeepSeek-R1 decreases from 74.37 \% to 62.44 \% at 15\%, retaining $\sim$84.0\% of baseline. These results fall outside the 90--100\% baseline retention regime observed on many non-coding tasks. 

We hypothesize that code debugging often depends on fine-grained syntactic and semantic dependencies that may be distributed across the repository; removing seemingly low saliency spans can break the local context required to pinpoint the injected error. Code debugging is less tolerant to extreme keep rates, and stronger structural constraints (e.g., dependency-aware span selection or minimum coverage of function-level context) may be required to maintain accuracy under heavy compression. We conclude that our current cross model speculative prefill framework is structure-dependent.  These findings indicate that while cross-family speculative prefill is a general and effective prompt compression primitive, extending it with task-aware structural constraints is a promising direction for future work.

\subsection{Qualitative Examples}
\label{app:qualitative_study}

Below we show a representative code debugging prompt after compression, using LLaMA-3.1-8B-Instruct as the draft model and DeepSeek-R1 as the target model. The keep rate is set to 30\%. Delimiter tokens ``// omitted'' are highlighted in red to indicate breaks between non-contiguous text spans. The ground-truth answer is highlighted in blue, and the final question is placed at the end of the prompt. The full compressed prompt is approximately 33k tokens long; for clarity, we show only selected excerpts below.

\noindent\rule{\linewidth}{0.4pt}

There is ONLY ONE function in the large project that is deliberately made to include an obvious error.
Please find the function that contains the most obvious errors. I will give you four options to narrow your scope.
You can inspect the options and think. Eventually, tell me the answer using one single letter (A, B, C, or D).

\bigskip
\textbf{Package:} \texttt{pyarmor}

\bigskip
\textbf{File:} \texttt{pyarmor/cli/core/features.py}

\begin{Verbatim}[fontsize=\footnotesize, commandchars=\\\{\}]
#! /usr/bin/env python
# -*- coding: utf-8 -*-

# Each log
#    revision, age, (new features), (changed features), (removed features)
__CHANGE_LOGS__ = (
    (1, 0, (), (), ()),
)

class PyarmorFeature(object):

    def features(self):
        '''return features list from change logs'''
        result = set()
        [result.update(item[2]) for item in __CHANGE_LOGS__]
        return result

\end{Verbatim}

\bigskip
\textbf{File:} \texttt{pyarmor/cli/core/\_\_init\_\_.py}

\begin{Verbatim}[fontsize=\footnotesize, commandchars=\\\{\}]
#! /usr/bin/env python
# -*- coding: utf-8 -*-
#
__VERSION__ = '5.4.0'

def format_platform():
    import platform
    import sys
    from struct import calcsize

    def format_system():
        plat = platform.system().lower()
        plat = ('windows' if plat.startswith('cygwin') else
                'linux' if plat.startswith('linux') else
                'freebsd' if plat.startswith(
                    ('freebsd', 'openbsd', 'isilon onefs')) else plat)
        if plat == 'linux
\omitted
        
class PyarmorRuntime(object):

    @staticmethod
    def get(plat, extra=None, native=True):
        from os import scandir, path as os_path
        prefix = 'pyarmor_runtime'

        # Themida is only available for windows
        if extra == 'themida' and not plat.startswith('windows'):
            extra = None

        pkgpath = os_path.dirname(__file__)
        if native and not extra:
            path = pkgpath
            for entry in scandir(path):
                parts = entry.name.split('.')
                if parts[0] == prefix and parts[-1] in ('so', 'pyd', 'dylib'):
                    return entry.name, os_path.abspath(entry.path)

        dirnames = map_platform(plat).split('.')
        path = os_path.join(pkgpath, extra if extra else '', *dirnames)

\omitted
 getLogger('cli').info('fallback to pyarmor.cli.runtime==%s', ver)
            return PyarmorRuntime.get(plat, extra=extra)
        except ModuleNotFoundError:
            pass

\bughl{def repack_carchive(executable, pkgfile, buildpath, obfpath, rtentry):}
\bughl{    pkgarch = CArchiveReader2(executable}
\omitted
\bughl{.dirname(executable), rtname)}
\bughl{            os.makedirs(dest, exist_ok=True)}
\bughl{            shutil.copy2(rtbinary, dest)}

\bughl{        repack_executable(executable, buildpath, obfpath, rtentry, codesign)}

\bughl{    def _fixup_darwin_rtbinary(self, rtbinary, pylib_name):}
\bughl{        '''Unused since Pyarmor 8.3.0'''}
\bughl{        from sys import version_info as pyver}
\bughl{        pylib = os.path.normpath(os.path.join('@rpath', pylib_name))}
\bughl{        output = check_output(['otool', '-L', rtbinary])}
\bughl{        for line in}
\omitted
\bughl{    main_entry()}

@classmethod
def build_globfiles(cls, patterns, path=''):
    files = []
    n = len(path) + 1
    for x in patterns:
        for name in glob(os.path.join(path, x)):
            files.append(name[n:])
    return set(files)

def info(self):
    lines = []
    for k, v in Project.DEFAULT_VALUE:
        if k == 'build_time':
            v = time.asctime(time.gmtime(self[k]))
        else:
            v = str(self[k])
            n = 50
            if len(v) > n:
                v = v[:n] + '\textbackslash n \%24s'\%'' + v[n:]
        lines.append('\%22s: \%s' % (k, v))
    return '\textbackslash n'.join(lines)

if __name__ == '__main__':
    project = Project()
    
\end{Verbatim}

\bigskip
Which function has deliberate error?
\begin{itemize}
\item \textbf{A.} Resource.pkgname
\item \textbf{B.} repack\_carchive
\item \textbf{C.} cmd\_gen
\item \textbf{D.} \_\_init\_\_
\end{itemize}

You should first find the functions in the options. Repeat their content, inspect through code,
and at last give me your answer for the function that has the deliberate and obvious error in A, B, C, or D.

\end{document}